\documentclass{article} 
\usepackage{iclr2016_conference,times}
\usepackage{hyperref}
\usepackage{url}
\usepackage{times}
\usepackage{epsfig}
\usepackage{graphicx}
\usepackage{amsmath}
\usepackage{amssymb}
\usepackage{listings}
\DeclareMathAlphabet{\pazocal}{OMS}{zplm}{m}{n}

\usepackage{expl3}
\ExplSyntaxOn
\newcommand\latinabbrev[1]{
  \peek_meaning:NTF . {
    #1\@}%
  { \peek_catcode:NTF a {
      #1.\@ }%
    {#1.\@}}}
\ExplSyntaxOff

\def\eg{\latinabbrev{e.g}}

\def\ie{\latinabbrev{i.e}}

\title{Learning Dense Convolutional Embeddings for Semantic Segmentation}

\author{Adam W.~Harley \& Konstantinos G.~Derpanis\\
Ryerson University\\
\texttt{\{aharley,kosta\}@scs.ryerson.ca} \\
\And
Iasonas Kokkinos \\
CentraleSup\'elec and INRIA \\
\texttt{iasonas.kokkinos@ecp.fr} \\
}

\begin{document}

\maketitle

\begin{abstract}This paper proposes a new deep convolutional neural network (DCNN) architecture that learns pixel embeddings, such that pairwise distances between the embeddings can be used to infer whether or not the pixels lie on the same region. That is, for any two pixels on the same object, the embeddings are trained to be similar; for any pair that straddles an object boundary, the embeddings are trained to be dissimilar. Experimental results show that when this embedding network is used in conjunction with a DCNN trained on semantic segmentation, there is a systematic improvement in per-pixel classification accuracy. Our contributions are integrated in the popular Caffe deep learning framework, and consist in straightforward modifications to convolution routines. As such, they can be exploited for any task involving convolution layers.
\end{abstract}

\section{Introduction}

Deep convolutional neural networks (DCNNs) \citep{lecun-98} are the method of choice for a variety of high-level vision tasks \citep{astounding}. Fully-convolutional DCNNs have recently been a popular approach to semantic segmentation, because they can be efficiently trained end-to-end for pixel-level classification \citep{sermanet-iclr-14,chen_deeplab,long_shelhamer_fcn}. 

A weakness of DCNNs is that they tend to produce smooth and low-resolution predictions, partly due to the subsampling that is a result of cascaded convolution and max-pooling layers. Many different strategies have been explored for sharpening the boundaries of predictions produced by fully-convolutional DCNNs. One popular strategy is to add a dense conditional random field (CRF) to the end of the DCNN, introducing contextual information to the segmentation via long-range dependencies in the CRF \citep{chen_deeplab, lin_piecewise}. Another strategy is to reduce the subsampling effected by convolution and pooling, by using the ``hole'' algorithm for convolution \citep{chen_deeplab}. A third strategy is to add trainable up-sampling stages to the network via ``de-convolution'' layers in the DCNN \citep{noh2015learning,long_shelhamer_fcn}. 

\begin{figure}[t]
\begin{center}
   \includegraphics[trim={3cm 3.5cm 8cm 12.5cm},clip,width=0.9\linewidth]{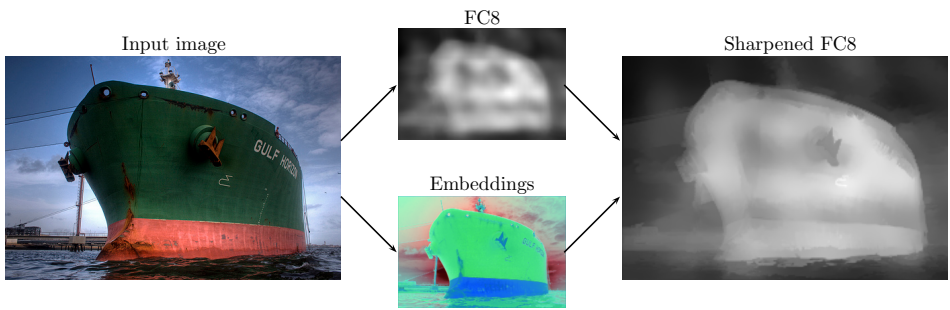}
  
\end{center}
\vspace{-6pt}
\caption{Semantic segmentation predictions produced by DCNNs can be sharpened by filtering, via this work's learned embeddings trained to capture semantic region similarity.}
\label{fig:boat}
\vspace{-12pt}
\end{figure}

This paper's strategy, which is complementary to those previously explored, is to train the network to produce segmentation-like internal representations, so that foreground pixels and background pixels within local patches can be treated differently. In particular, the aim is to increase the sharpness of the DCNN's final output by using local pixel affinities to filter and re-weight the final layer's activations. For instance, as can be seen in Figure~\ref{fig:boat}, if a DCNN is centered on a ``boat'' pixel, but the surrounding patch includes some pixels from an occluder or the background, the DCNN's final prediction will typically reflect the presence of the distractors by outputting a mix of ``boat'' and ``background''. The approach of this paper is to learn and use semantic affinities between pixels, so that the DCNN output centered at the ``boat'' pixel can be strengthened by using information from other ``boat'' pixels within the patch. More generally, the approach allows the prediction at any pixel to be replaced with a weighted average of the similar neighboring predictions. This has the effect of sharpening the predictions at object boundaries, while making predictions within object boundaries more uniform. 

The key to accomplishing this is to have the network produce internal representations that lend themselves to pairwise comparisons, such that any pair that lies on the same object will produce a high affinity measure, and pairs that straddle a boundary produce a low affinity measure. Prior work has investigated the use of affinity cues in similar contexts \citep{ren2003learning,dai2014convolutional}, but these required handcrafted algorithms for computing the affinity information, and would typically be pre-computed in a separate process. The current work is unique for learning the cues directly from image data, and for computing the affinities densely and ``on the fly'' within a DCNN. 

The learned embeddings and their distance functions are implemented efficiently as convolution-like layers in Caffe \citep{jia2014caffe}. The embedding layers can either be trained independently, or integrated in the full DCNN pipeline and trained end-to-end (with or without per-pixel labels). Source code and trained embeddings will be publicly released.

\section{Related work}

This work is closely related to three major research topics in feature learning and computer vision: metric learning, segmentation-aware descriptor construction, and DCNNs.

\textbf{Metric learning.} The goal of metric learning is to produce features from which one can estimate similarity between pixels or regions in the input \citep{frome2007learning}. \citet{siamese} influentially began learning these descriptors in a convolutional network, and subsequent related work has yielded compelling results for tasks such as wide-baseline stereo correspondence \citep{han2015matchnet, ZagoruykoCVPR2015, vzbontar2014computing}. Recently, the topic of metric learning has been studied extensively in conjunction with image descriptors, such as SIFT and SID \citep{trulls2013dense, simo2015discriminative}, improving the applicability of those descriptors to patch-matching problems. Most prior work in metric learning has concerned the task of finding one-to-one correspondences between pixels seen from different viewpoints. The current work, in contrast, concerns the task of matching all pairs of points that lie on the same region. This requires a higher degree of invariance than has previously been necessary -- not only to rotation, scale, and partial occlusion, but to objects' interior details. 

\textbf{Segmentation-aware descriptors.} The purpose of a segmentation-aware descriptor is to capture the appearance of the foreground while being invariant to changes in the background or occlusions. To date, most work in this domain has been on developing handcrafted segmentation-aware descriptors. For instance, soft segmentation masks \citep{ott2009implicit,leordeanu2012efficient} and boundary cues \citep{maire2008using,shi2000normalized} have been used to augment features like SIFT and HOG, to suppress contributions from pixels likely to come from the background \citep{trulls2013dense,trulls2014segmentation}. The intervening contours algorithm \citep{fowlkes2003learning} provides another type of affinity measure, used previously for image segmentation. The current work's ``embeddings'' are a first attempt at developing fully-learned segmentation-aware descriptors. As a secondary contribution, the current work also implements intervening contours as a layer in a DCNN, using deep-learned boundary cues from another work \cite{xie15hed}. Since the boundary cues require a separate DCNN, this is meant to represent a costly alternative to the learned embeddings featured here.

\textbf{DCNNs for semantic segmentation.} Fully-convolutional DCNNs are fast and effective semantic segmentation systems \citep{long_shelhamer_fcn}. Part of the appeal of DCNNs is they can be trained end-to-end, without the need for any handcrafted feature representations. However, DCNNs' repeated subsampling (through strided convolution and max-pooling) presents an issue, because it reduces the resolution at which the DCNN can make predictions. This has been partially addressed by trainable up-sampling layers \citep{long_shelhamer_fcn}, and the ``hole'' algorithm for convolution \citet{chen_deeplab}, but state-of-the-art systems also attach a dense CRF \cite{koltun2011efficient} to the DCNN  to increase the sharpness of the output. The CRF can be trained as a separate module \citep{chen_deeplab}, or jointly with the DCNN \citep{lin_piecewise}, though both cases are at significant added computational cost. Another approach to the subsampling issue, more in line with the current paper, is to incorporate segmentation cues into the DCNN. \citet{dai2014convolutional} recently used superpixels to generate masks for convolutional feature maps, enforcing sharp contours in their outputs. The current paper takes this idea further, by replacing the sparse handcrafted segmentation cues with dense learnable variants.

\textbf{Contributions.} In the light of the related work, this paper's main contributions are as follows. First, the paper uses a DCNN architecture to learn  pixel embeddings, such that pairwise distances between the embeddings indicate whether or not the pixels belong to the same region. Second, these embeddings (and their distance functions) are implemented as convolution-like layers in Caffe, with minimal computational overhead. Third, the learned embeddings are integrated with the state-of-the-art DeepLab semantic segmentation system, and this is shown to improve performance on the VOC2012 segmentation task.

\section{Technical approach}

This section establishes (1) how segmentation embeddings can be learned from pixel-wise labels, (2) how the embeddings can be merged with convolution, such that they can be learned without pixel-wise labels, and finally (3) how contour cues can be used to create an alternative affinity measure.

\subsection{Learning segmentation embeddings}

The goal of the current work is to train a set of convolutional layers to create dense ``embeddings'', which can be used to calculate pixel affinities relating to the semantic similarity of the underlying regions. Pixels pairs that share a semantic category should produce similar embeddings (\ie, a high affinity), and pairs that do not share a semantic category should produce dissimilar embeddings (\ie, a low affinity). 

This goal is represented in a loss function $\pazocal{L}$, which accumulates the quality of embedding pairs sampled across the image. In this work, pairwise comparisons are made between each pixel $i$ and its spatial neighbours $j \in N(i)$. Collecting pairs within a fixed window lends simplicity and tractability, although in general the pairs can be collected at any range. Denoting the quality of a particular pair of embeddings with $\ell_{ij}$, the overall loss is defined as 
\begin{equation} \label{eq:loss}
\pazocal{L} = \sum_{i\in I} \sum_{j \in N(i)} \ell_{ij}.
\end{equation}
The network is trained to minimize this loss through stochastic gradient descent. 

The inner loss function $\ell_{ij}$ represents how well a pair of embeddings $e_i$ and $e_j$ respect the affinity goal. Pixel-wise labels are a convenient resource for quantifying this loss, since they can provide information on whether or not the pixels belong to the same region. Using this information, the distance between embeddings can be optimized according to label parity. That is, same-label pairs can be optimized to have a small distance, and different-label pairs can be optimized to have a large distance. Denoting the label of pixel $i$ with $\l_i$, and the embedding at that pixel with $e_i$, the inner loss is defined as 
\begin{equation} \label{eq:miniloss}
    \ell_{ij} =  
    \bigg\{
    \begin{array}{lr}
      \text{max} \left( |e_i - e_j| - \alpha, 0 \right)  & \text{if } l_i=l_j, \\
      \text{max} \left( \beta - |e_i - e_j|, 0 \right)  & \text{if } l_i \neq l_j, \\
    \end{array}
\end{equation}
where $\alpha$ and $\beta$ are design parameters that specify the ``near'' and ``far'' thresholds against which the embedding distances are compared. In this work, $\alpha = 0.5$, and $\beta = 2$ are used. 

\begin{figure}[t]
\begin{center}
   \includegraphics[trim={3.1cm 4.0cm 6.7cm 9.6cm},clip,width=0.9\linewidth]{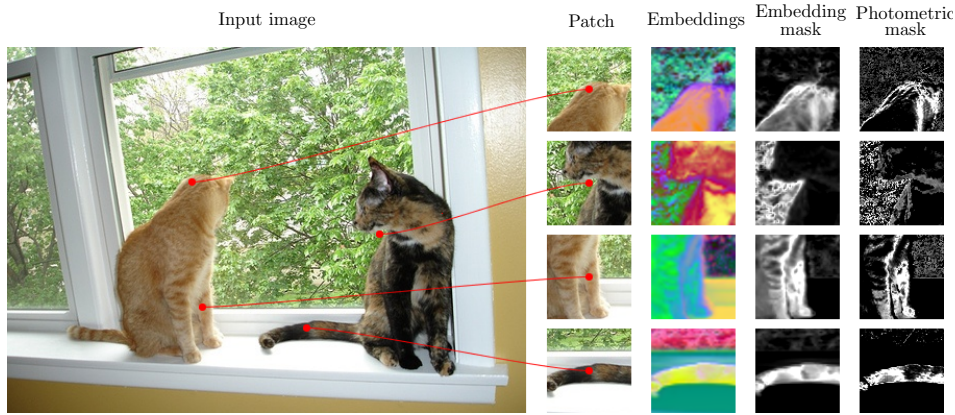}
\end{center}
\vspace{-6pt}
   \caption{Embeddings and local masks are computed densely for input images. For four locations in the image shown on the left, the figure shows the extracted patch, embeddings (compressed to 3 dimensions by PCA, for visualization purposes), and embedding-based mask (left-to-right). For comparison, the mask generated by photometric color distances is shown on the far right. }
\label{fig:rgbvs}
\vspace{-12pt}
\end{figure}

The embedding distances can be computed with any distance function. In this work, $L_1$ and $L_2$ norms were tried. Embeddings learned from both distances produced visually appealing masks, but the $L_1$-based embeddings were found to be easier to train. Specifically, it was found that the $L_1$-based embeddings can safely be trained with a higher learning rate than $L_2$-based ones, because they are less vulnerable to the problem of exploding gradients. Figure~\ref{fig:rgbvs} shows visualizations of the $L_1$-based embeddings learned by the network.

\subsection{Embedding-based segmentation masks}

Once these embeddings are learned, they can be used to create segmentation masks. For a pixel $i$ and a neighbour pixel $j \in N(i)$, one can define 
\begin{equation} \label{eq:pixelmask}
m_i = \exp ( - \lambda | v_i - v_j | )
\end{equation}
to be the weight applied to pixel $j$ in a mask centered on $i$, where $\lambda$ is a parameter specifying the hardness of the mask. This parameter can be learned inside a DCNN. The exponential function scales the masks to the range $[0,1]$, where similar pixels are given values near $1$, and dissimilar pixels are given values near $0$. 

These masks can be applied convolutionally, so that the output at location $i$ becomes 
\begin{equation} \label{eq:mask}
y_i = \sum_{j \in N(i)} m_j x_j,
\end{equation}
where $x_j$ is the input at location $j$. If $x_j$ is a vector, the mask is simply applied to every element of the vector. The effect of this is to replace each input with a weighted average of its similar neighbours. Since the affinities capture semantic similarity, this is expected to improve the quality of the output. 

Finally, each output is normalized the sum total of the mask, so that the output magnitudes do not change as a function of neighbourhood size. With normalization, the masking equation~\eqref{eq:mask} becomes

\begin{equation} \label{eq:normalizedmask}
y_i = \frac{\sum_{j \in N(i)} m_j x_j}{\sum_{j \in N(i)} m_j}.
\end{equation}

Note that if $x_j$ is an RGB value, and $m_j$ is a Gaussian that jointly captures RGB and geometric distance between pixels $i$ and $j$, the masking equation~\eqref{eq:mask} is equivalent to the bilateral filter \citep{tomasi1998bilateral}, which is a well-known technique in signal processing for smoothing while preserving edges. Since the filter in the current work depends on the embeddings, and the embeddings are learned in a DCNN, the current approach represents a generalization of the bilateral filter. Interestingly, the \cite{koltun2011efficient} algorithm for dense CRFs is also related to the bilateral filter, in the sense that the inference step, through mean field approximation, essentially accomplishes a repeated application of a non-linear filter. This shared connection is appropriate, since CRFs and embedding-based segmentation masks have common goals: to sharpen predictions at object boundaries while smoothing in the interior.

When the embeddings are integrated into a larger network that uses them for masks, the embedding loss function~\eqref{eq:loss} is no longer necessary. Since all terms of the normalized masking equation~\eqref{eq:normalizedmask} are differentiable, the global objective (\eg, classification accuracy) can be used to tune not only the input term $x_j$, but also the mask term $m_j$. Therefore, the embeddings can be learned end-to-end in the network when used to create masks. 

In this work, the embeddings are trained first with a dedicated loss, then fine-tuned in the larger pipeline as masks. Figure~\ref{fig:rgbvs} shows examples of the learned embeddings and masks, as compared with masks created by photometric color distances. 

\subsection{Contour-based affinities}

This work also explores the use of contour cues to generate pixel affinities. In particular, the interest is to see how contour-based affinities compare with the embedding-based affinities. Automatic boundary detection with DCNNs has recently attained excellent results \citep{xie15hed}, so contour-based affinity cues should be a strong baseline to compare against. 

For contour cues, this work uses the learned contour cues of a state-of-the-art DCNN trained for the task, named the Holistically-nested Edge Detection (HED) network \citep{xie15hed}. For each pixel $i$, the intervening contours algorithm \citep{fowlkes2003learning} is computed for each of its neighbours $j \in N(i)$, to determine the maximum probability of a contour being on a line that travels from $i$ to $j$. If two pixels $i$ and $j$ lie on different objects, there is likely to be a boundary separating them; the intervening contours step returns the probability of that boundary, as provided by the boundary cue. As with the embeddings, this step is implemented entirely within the DCNN. Intervening contours are computed with a specialized layer for the task. 

\section{Network architecture and implementation details}

This section first describes how the ideas of the technical approach were integrated in a DCNN architecture, and then establishes details on how the individual components were implemented efficiently as convolution-like DCNN layers.

\subsection{Network architecture}

\begin{figure}[t]
\begin{center}
   \includegraphics[trim={1.2cm 20.8cm 2.2cm 3.4cm},clip,width=0.8\linewidth]{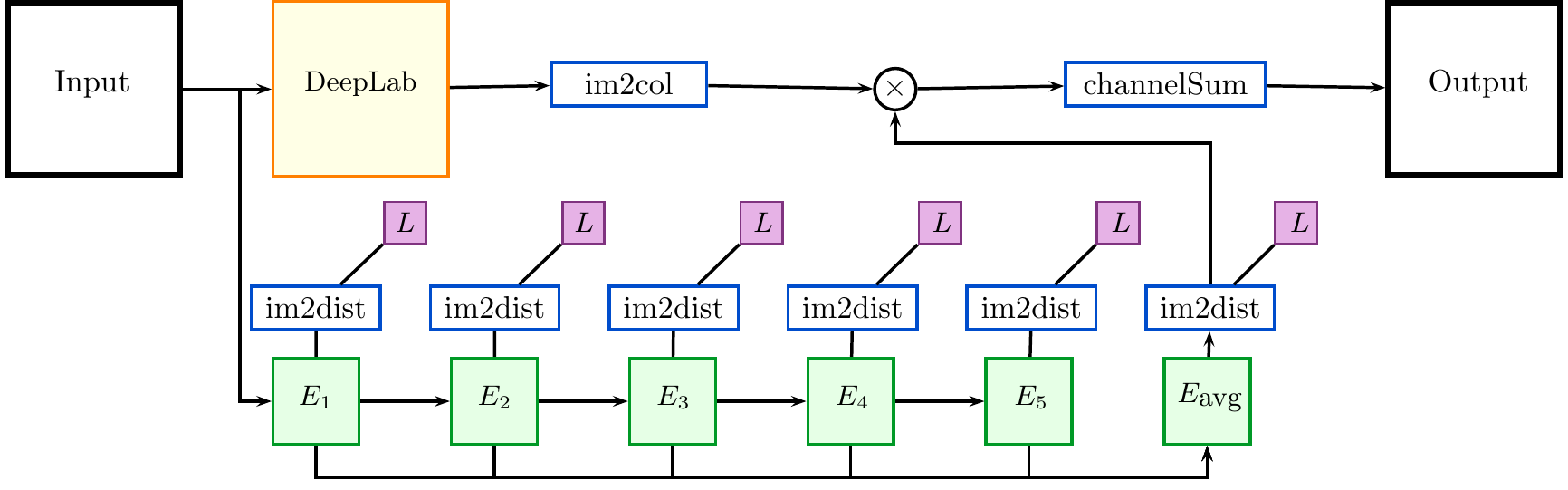}
\end{center}
\vspace{-12pt}
   \caption{Schematic for the DCNN featured in this work. The embedding layers (bottom half of the schematic) are the main contribution. Embedding layers are indicated with boxes labelled $E$; the final embedding layer creates a weighted average of the other embeddings. Loss layers, indicated with $\pazocal{L}$ boxes, provide gradients for each embedding layer.}
\label{fig:schematic}
\vspace{-12pt}
\end{figure}

Figure~\ref{fig:schematic} illustrates the full network featured in this paper. The input image is sent to two parallel processing streams, which merge later: a DeepLab network \citep{chen_deeplab}, and an embeddings network. Both networks are modelled after the VGG-16 network \citep{Chatfield14}. The DeepLab network is the multi-scale large field-of-view model from \cite{chen_deeplab}. 

The embeddings network has the following design. The first five layers are initialized from the earliest convolutional layers in VGG-16. There is a pooling layer after the second layer, and after the fourth layer, so the five layers capture information at different scales. The output from each of these layers is sent to pairwise distance computations ($im2dist$) followed by a loss, so that each layer develops embedding-like representations. The idea of using a loss at each intermediate layer is inspired by \cite{xie15hed}, who used this strategy to learn boundary cues in a DCNN. 

The outputs from the intermediate embedding layers are upsampled to a common resolution, concatenated, and sent to a randomly-initialized convolutional layer with $1 \times 1$ filters and $64$ outputs. This layer learns a weighted average of the first five convolutional layers' outputs, and creates the final 64-dimensional embeddings. The output of this layer trained in the same way as the others (with $im2dist$ and a loss), and is used as the final embeddings. The final embeddings are used to mask the output of DeepLab's final convolutional layer (\ie, ``fc-fusion''), and then sent to a softmax layer to form prediction scores. 

To allow an evaluation that treats the embeddings as a modular upgrade to existing semantic segmentation systems, the DeepLab network was kept as a fixed component, and never fine-tuned with the masks. Better performance can be achieved by training the full pipeline end-to-end. 

Although the size of the window used for computing embedding distances has an important effect at test time (since it specifies the radius in which to search for contributing information), the window size was not found to have a substantial effect at training time. The embeddings used in the experiments were trained with losses computed in $9 \times 9$ windows, with a stride of $2$. 

To implement the intervening contours approach to pixel affinities, the state-of-the-art HED network \cite{xie15hed} was used to compute boundary cues, and affinities were computed with the intervening contours algorithm. Cross-validation with a step size of 5 on the hardness parameter $\lambda$ led to the choice of $\lambda=5$. Since the HED network has considerable memory requirements, achieving results directly comparable to those computed with embedding-based affinities is not actually feasible in the memory constraints of a Tesla K-40 GPU. To overcome this constraint, boundary cues were pre-computed for the entire PASCAL VOC validation set.

\subsection{Efficient convolutional implementation details}

This section provides the implementation details that were required to efficiently integrate the embeddings, masks, and intervening contours, with DCNNs. Source code for this work will be made available online. All new layers are implemented both for CPU and GPU, and are as fast as \textit{im2col}.

Computing pairwise distances densely across the image is a computationally expensive process. The current work implements this efficiently by solving it in the same way Caffe \citep{jia2014caffe} realizes convolution: via an image-to-column (\textit{im2col}) transformation, followed by matrix multiplication. 

The current work implements dense local distance computation in a new DCNN layer named \textit{im2dist}. For every position $i$ in the feature-map provided by the layer below, a patch of features is extracted from the neighborhood $j \in N(i)$, and local distances are computed between the central feature and its neighbours. These distances are arranged into a column vector of length $K$, where $K$ is the total dimensionality of a patch. This process turns an $H \times W$ feature-map into an $H \times W \times K$ matrix, where each element in the $K$ dimension holds a distance. 

To turn these distances into masks, the matrix is passed through an exponential function with a particular hardness. This corresponds to the mask term definition~\eqref{eq:pixelmask}, where the hardness parameter is specified by $\lambda$. In this work, $\lambda = 30$ was chosen, based on cross-validation with a step size of $5$. 

To perform the actual masking, the input to be masked must simply be processed by $im2col$ (producing another $H \times W \times K$ matrix), then multiplied pointwise with the masking matrix, and summed across $K$. This accomplishes the masking equation~\eqref{eq:mask}. 

The resulting matrix of predictions can optionally be normalized. To create the normalizing coefficients, \ie, the denominator in the normalized masking equation~\eqref{eq:normalizedmask}, the masking matrix must simply be summed across $K$ to create a mask sum for every location. The masked output can then be pointwise divided with the mask sums, creating the final normalized masked output.                                                 

The loss function for the embeddings~\eqref{eq:loss} is implemented using similar computational techniques. First, the label image is processed with a new layer named $im2parity$. This creates an $H \times W \times K$ matrix in which for each pixel $i$, the $K$-dimensional column specifies (with $\{0,1\}$) whether or not the local neighbours $j \in N(i)$ have the same label. The result of this process can then be straightforwardly combined with the result of $im2dist$, to penalize each distance according to the correct loss case and threshold in the pairwise loss equation~\eqref{eq:miniloss}.

Finally, the intervening contours algorithm is implemented in a similar way, in a layer named \textit{im2interv}. For every position $i$ in a boundary probability map provided by the layer below, a $K$-dimensional column vector is generated, representing the intervening contour output for each position in a neighborhood centered on $i$. Specifically, for each position in this neighborhood, a straight line is traced from that position to the center position (using the \cite{bresenham1965algorithm} algorithm), and the maximum boundary probability along that line is stored. The $H \times W \times K$ output of this process can be used in exactly the same way as the output of $im2dist$.

\section{Evaluation}

The baseline for the evaluation is the current best publicly-released DeepLab network (``Deep-MSc-Coco-LargeFOV''; \cite{chen_deeplab}), which is a strong baseline for semantic segmentation. This model was initialized from a VGG network trained on ImageNet, then trained on the Microsoft COCO training and validation sets \citep{coco}, and finally fine-tuned on training and validation sets of the PASCAL VOC 2012 challenge \citep{pascal-voc-2012}. This network is augmented with embeddings learned on the COCO dataset. Additional baselines are provided by RGB distances, pre-computed state-of-the-art handcrafted \cite{leordeanu2012efficient} 8-dimensional embeddings (with default parameters), and pre-computed intervening contour affinities. All baselines were tested using a $9 \times 9$ affinity window (applied as mask once).

Evaluation on the PASCAL VOC validation set is presented in Table~\ref{table:val}. Note that the publicly-released DeepLab model was trained on the VOC trainval set, so results on the validation set cannot be understood to reflect performance at test time, but rather reflect training error. However, since the embeddings were not trained on any VOC data, improvements to this performance do reflect general improvements in the model. Accordingly, the experiments of this paper show that improvements on the validation set translate to improvements on the test set.

The results first of all show that using learned embeddings to mask the output of DeepLab systematically provides a 0.5\% to 1.5\% improvement in mean intersection-over-union (IOU) accuracy. , the improvements exceed those attainable by RGB, intervening contours, or handcrafted \cite{leordeanu2012efficient} embeddings. While the numerical difference between the performance of these affinity cues is small (\eg, there is a 0.03\% difference between handcrafted embeddings and learned embeddings when both are computed in a $9 \times 9$ and applied once), it is important to emphasize that the learned embeddings are uniquely capable of being fine-tuned. That is, this initial comparison uses the learned embeddings ``off-the-shelf'' (as they were learned from the COCO dataset), but realistic use would involve fine-tuning on all available data. The current work's experiments on the VOC test set benefit from this fine-tuning.

Additional validation experiments reported in Table~\ref{table:val} explore the effects of two design critical parameters: filter window size, and the number of times to apply the filter. A wider window, though more expensive, is expected to improve performance by allowing information from a wider radius to contribute to each prediction. The approach was evaluated at window sizes of $7 \times 7$, $9 \times 9$, $11 \times 11$, and finally $15 \times 15$. The results confirm that increasing the filter size improves performance. 

The second design parameter is the number of times to apply the filter. Once the embeddings and masks are computed, it is trivial to run the masking process repeatedly. Applying the process multiple times is expected to improve performance, by strengthening the contribution from similar neighbours in the radius. This second parameter also effectively increases the number of contributing neighbours: after filtering with $7 \times 7$ filters, each new $7 \times 7$ region effectively contains information from an $11 \times 11$ area in the original input. Therefore, along with results from applying each filter once, results are presented for applying each filter the maximum possible number of times, given the memory constraints of a Tesla K-40 GPU. As expected, repeating the application of the filter improves performance. The best model applied a $9\times9$ mask seven times recursively.

The best embedding configuration determined from the PASCAL VOC validation set was fine-tuned end-to-end in its full pipeline on the VOC trainval set, and submitted to the VOC test server. As shown in Table~\ref{table:test}, improvement on the test set was approximately 1.2\% over the baseline, roughly consistent with the results observed on the validation set.

\begin{figure}[t]
\begin{center}
   \includegraphics[trim={3cm 3cm 4.4cm 5.6cm},clip,width=0.9\linewidth]{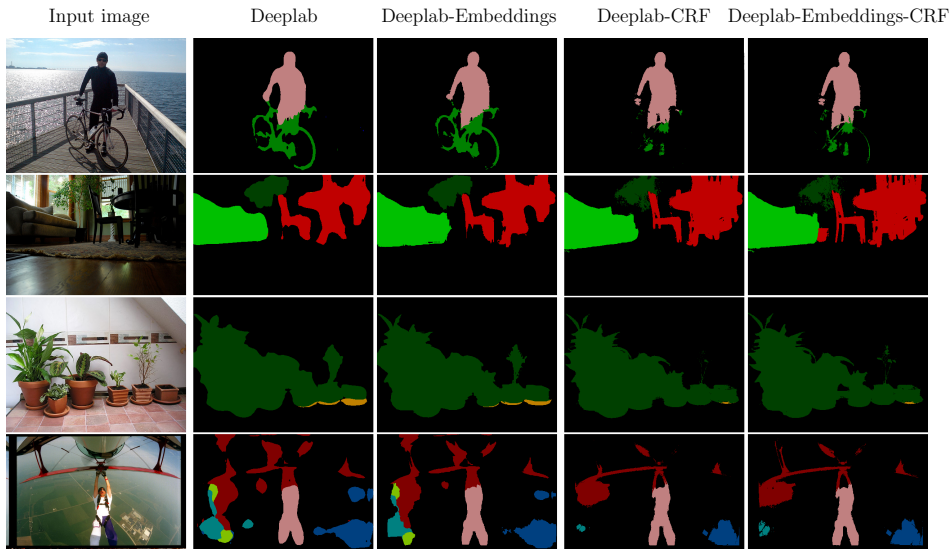}
   \vspace{-12pt}
\end{center}
\vspace{-12pt}
   \caption{Visualizations of the semantic segmentations produced by various considered approaches. Input is shown at far left, followed by (left-to-right) DeepLab, DeepLab with $9 \times 9$ embeddings (applied seven times), DeepLab with a dense CRF, and DeepLab with embeddings and a dense CRF.}
\label{fig:outputs}
\vspace{-12pt}
\end{figure}

As discussed earlier, dense CRFs are often used to sharpen the predictions produced by DCNN-based semantic segmentation systems like DeepLab. To test if the improvement offered by embedding-based masks continues through the application of the CRF, a dense CRF \citep{koltun2011efficient} was trained on top of the mask-sharpened outputs, using the PASCAL VOC validation set for cross-validation of its hyper-parameters. As shown in Table~\ref{table:test}, the embedding-augmented DeepLab outperforms the DeepLab-CRF baseline by 0.4\%. Visualizations of the results are shown in Figure~\ref{fig:outputs}.

\begin{table}
\centering
\begin{minipage}{.48\textwidth}
\caption{VOC 2012 validation results (``IC'' indicates intervening contour-based filters; ``Leo'' indicates \cite{leordeanu2012efficient} embeddings).}
\begin{center}
\begin{tabular}{l|c}
\textbf{Method} & \textbf{IOU (\%) }\\
\hline
DeepLab & 78.22 \\
\hline
DeepLab, $9 \times 9$ RGB x1 & 78.30\\
DeepLab, $9 \times 9$ Leo x1 & 78.86\\
DeepLab, $9 \times 9$ IC x1 & 78.85\\
\hline
DeepLab, $7 \times 7$ embeds x1 & 78.70\\
DeepLab, $7 \times 7$ embeds x12 & 79.68\\
DeepLab, $9 \times 9$ embeds x1 & 78.89\\
DeepLab, $9 \times 9$ embeds x7 & \textbf{79.69}\\
DeepLab, $11 \times 11$ embeds x1 & 79.03 \\
DeepLab, $11 \times 11$ embeds x4 & 79.64\\
DeepLab, $15 \times 15$ embeds x1 & 79.39 \\
\end{tabular}
\end{center}
\label{table:val}
\end{minipage}\hfill
\begin{minipage}{.48\textwidth}
\caption{VOC 2012 test results.}
\begin{center}
\begin{tabular}{l|c}
\textbf{Method} & \textbf{IOU (\%)} \\
\hline
DeepLab & 70.31 \\
DeepLab, $9 \times 9$ embeds x7 & \textbf{71.54} \\
\hline
DeepLab-CRF & 73.60 \\
DeepLab-CRF, $9 \times 9$ embeds x7 & \textbf{74.00} \\
\end{tabular}
\end{center}
\label{table:test}
\end{minipage}\hfill
\end{table}

\section{Conclusion}
This paper proposed a new deep  convolutional neural network architecture for learning embeddings. Results showed that integrating the embeddings into a strong baseline DCNN systematically improved results by a noticeable margin on both validation and testing in the PASCAL VOC 2012 dataset. Compared to results achieved through RGB distances, intervening-contour based affinities with state-of-the-art boundary cues, and state-of-the-art handcrafted embeddings, the performance of learned embeddings is higher, despite having a much smaller memory footprint. This approach to improvement is orthogonal to many others pursued in semantic segmentation, and it is implemented efficiently in the popular Caffe deep-learning framework, making it a useful and straightforward augmentation to existing semantic segmentation frameworks. Finally, although semantic segmentation is the targeted application of the current work, the overall approach does not depend on pixel-wise labels, and the embedding and masking layers can be used in any task involving DCNNs.

\bibliography{bibref_definitions_short,refs_short,iclr2016_conference}
\bibliographystyle{iclr2016_conference}

\end{document}